\documentclass[10pt,twocolumn,letterpaper]{article}

\usepackage[pagenumbers]{wacv} 

\usepackage{graphicx}
\usepackage{amsmath}
\usepackage{amssymb}
\usepackage{booktabs}

\usepackage{algorithm}
\usepackage{algpseudocode}
\usepackage{enumitem}
\usepackage{pifont}
\usepackage{makecell}
\usepackage{xcolor}
\usepackage{pifont}
\usepackage{multirow}
\usepackage[pagebackref,breaklinks,colorlinks]{hyperref}

\usepackage[capitalize]{cleveref}
\crefname{section}{Sec.}{Secs.}
\Crefname{section}{Section}{Sections}
\Crefname{table}{Table}{Tables}
\crefname{table}{Tab.}{Tabs.}

\def\wacvPaperID{2631} 
\def\confName{WACV}
\def\confYear{2025}

\begin{document}

\title{DG-MVP: 3D Domain Generalization via Multiple Views of Point Clouds for Classification}

\author{Huantao Ren\\
Syracuse University\\
{\tt\small hren11@syr.edu}
\and
Minmin Yang\\
Syracuse University\\
{\tt\small myang47@syr.edu}
\and
Senem Velipasalar\\
Syracuse University\\
{\tt\small svelipas@syr.edu}
}
\maketitle

\begin{abstract}
Deep neural networks have achieved significant success in 3D point cloud classification while relying on large-scale, annotated point cloud datasets, which are labor-intensive to build. 
Compared to capturing data with LiDAR sensors and then performing annotation, it is relatively easier to sample point clouds from CAD models. Yet, data sampled from CAD models is regular, and does not suffer from occlusion and missing points, which are very common for LiDAR data, creating a large domain shift. Therefore, it is critical to develop methods that can generalize well across different point cloud domains. 
Existing 3D domain generalization methods employ point-based backbones to extract point cloud features. Yet, by analyzing point utilization of point-based methods and observing the geometry of point clouds from different domains, we have found that a large number of point features are discarded by point-based methods through the max-pooling operation. 
This is a significant waste especially considering the fact that domain generalization is more challenging than supervised learning, and
point clouds are already affected by missing points and occlusion to begin with.
To address these issues, we propose a novel method for 3D point cloud domain generalization, which can generalize to unseen domains of point clouds.
Our proposed 
method employs multiple 2D projections of a 3D point cloud to alleviate the issue of missing points and involves a simple yet effective convolution-based model to extract features. The experiments, performed on the PointDA-10 and Sim-to-Real benchmarks, demonstrate the effectiveness of our proposed method, which outperforms different baselines, and can transfer well from synthetic domain to real-world domain.
\end{abstract}

\vspace{-0.6cm}
\section{Introduction}
\label{sec:intro}
With the rapid advancement and growing accessibility of 3D sensing technology, 3D point cloud analysis has garnered significant interest from the research community. 3D point cloud data is utilized in various applications, including self-driving cars, unmanned vehicles and robotics. 
\begin{figure}[t!]
\vspace{-0.2cm}
\begin{center}
\includegraphics[width=0.95\linewidth]{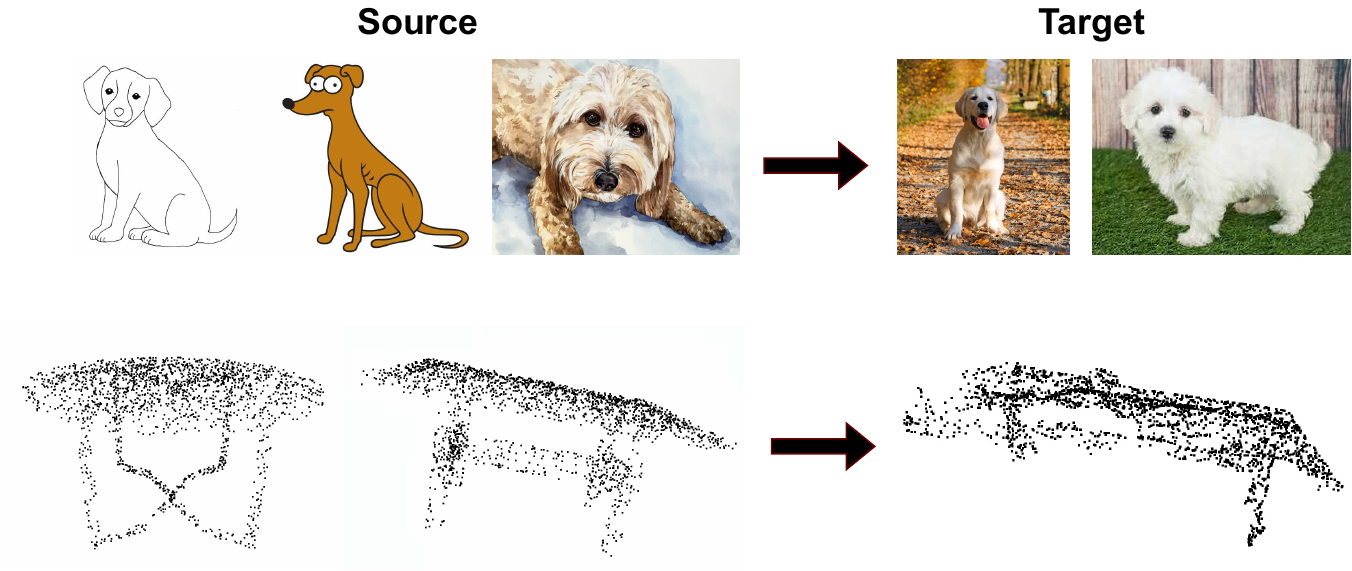}
\end{center}
\vspace{-0.4cm}
   \caption{Different domain shifts observed with 2D images and 3D point clouds.}
\label{fig:2Dvs3D}
\vspace{-0.7cm}
\end{figure}

In recent years, numerous deep neural network models~\cite{qi2017pointnet, feng2018gvcnn, qi2017pointnet++, li2018pointcnn, wang2019dynamic} have been proposed for 3D point cloud analysis. Although these methods have achieved impressive performance on  point cloud classification, they are trained and tested on the splits of the same dataset. However, for real-world applications, scanning objects with LiDAR sensors, and then performing annotation to build a sufficiently large and representative dataset is labor-expensive, and costly. Sampling point clouds from well-annotated CAD models, on the other hand, is relatively faster and easier. Thus, in practice, using a synthetic CAD-based dataset (e.g. ModelNet~\cite{wu20153d} and ShapeNet~\cite{chang2015shapenet}) as the training set and evaluating the model on a real-world dataset (e.g. ScanNet~\cite{dai2017scannet}) is a more labor-effective and efficient way. Yet, this approach introduces a significant domain shift challenge, since point clouds scanned from the real world often suffer from occlusion and missing points, whereas point clouds sampled from CAD models are very regular. The significant divergence between the distributions of the synthetic and real point clouds causes the performance of many existing point cloud classification models degrade significantly. Hence, it is essential to develop a 3D point cloud classification model to mitigate the domain shift issue.

In this paper, we focus on the 3D Domain Generalization (DG) problem. We train our proposed model, by only using synthetic point cloud data (referred to as the source data), which can then achieve high classification performance on real-world point cloud data (referred to as the target data) that is not accessible during training. With 2D images, 2D DG methods have been extensively investigated~\cite{volpi2019addressing, qiao2020learning, motiian2017unified, muandet2013domain, li2018learning}. Yet, the 
variations among 2D samples often stem from different styles, such as photo style, cartoon style and sketch. In contrast, the domain gap between 3D point cloud samples primarily arises from geometric differences, such as missing points and background noise, as shown in Fig~\ref{fig:2Dvs3D}. Thus, 2D DG methods cannot be readily applied to 3D data.

Existing approaches for 3D DG~\cite{huang2021metasets, huang2023sug, wei2022learning} use well-known point-based methods, such as DGCNN~\cite{wang2019dynamic} or PointNet~\cite{qi2017pointnet}, as their backbones due to their simplicity and effectiveness in representing 3D object shapes. PointNet employs multi-layer perceptron (MLP) and employs max-pooling operation at the end to extract permutation-invariant point features. DGCNN 
aggregates neighbors' features of each point in each Edge Convolution Layer, and also performs max-pooling at the end. In this work, we first show that point-based methods are not the most suitable models for 3D DG for the following reasons: (i) effectiveness of a point-based method in representing data is related to the number of points kept after max-pooling. Our experiments show that models with better DG ability tend to use more points for the final prediction. Considering that DG problem is already more challenging than supervised classification, it is even more important to make effective use of the available points. Many point-based methods employ max-pooling, 
which often leads to the loss of potentially useful points and reduces the shape representation ability; (ii) point-based methods are highly sensitive to occlusions and missing points. When generalized from synthetic to real-world domain, performance of these methods significantly degrades due to the simulation-to-reality gap.

To address the aforementioned issues, we propose an effective 3D domain generalization network, referred to as the DG-MVP, using multiple views of point clouds for point cloud classification. Instead of a point-based approach, our proposed DG-MVP employs multiple 2D projections of a point cloud as input.
We first project a 3D point cloud onto six orthogonal planes, by using SimpleView~\cite{goyal2021revisiting}, to obtain six depth images. Several example depth images obtained from different angles can be seen in Fig.~\ref{fig:projection}. We employ Resnet18~\cite{he2016deep} as the backbone to 
to transform each input depth image into a 3D feature map with the height, width and channel dimensions. Then, Depth Pooling is used across all six depth images, outputting the most prominent features from each depth image. Next, to learn more local features, we propose a multi-scale max-pooling module (MMP), which horizontally divides the feature map into several parts, and applies max-pooling on each part
to obtain the final representation. The effectiveness of our proposed method is analyzed through experiments, showing that it outperforms many point-based and multi view-based methods on DG of point cloud classification. To further improve the model robustness against geometry variations, we also employ two point transformation approaches proposed by Huang et al.~\cite{huang2021metasets} to simulate missing points and changes in scanning density during training. The experiments conducted on 3D domain adaptation and 3D DG benchmarks (PointDA-10~\cite{qin2019pointdan} and Sim-to-Real~\cite{huang2021metasets}) demonstrate that our DG-MVP consistently outperforms the state-of-the-art (SOTA) methods. 

Main contributions of this work include the following:\vspace{-0.2cm}
\begin{itemize} [leftmargin=0.4cm, itemsep=0.1pt] 
    \item We first analyze point utilization of the most commonly used point-based methods, and argue that they are not suitable for the 3D DG task. 
    \item We visualize projected depth images of point clouds, and observe that certain projections remain robust to missing points and deformations. Motivated by these observations, we propose DG-MVP, as an effective 3D domain generalization network using multiple views of point clouds for classification. 
    \item We propose a multi-scale max pooling module (MMP) to generate more local and descriptive features.
    \item Our proposed DG-MVP outperforms different baselines on PointDA-10 and Sim-to-Real datasets, even surpassing some 3D domain adaptation methods that utilize target data during training.
    \item We perform ablation studies to show the effectiveness of the components of the DG-MVP.
\end{itemize}

\section{Related Work}
\subsection{Point Cloud Classification}
Based on input format, point cloud classification methods can be classified into three categories, namely volumetric-based, multi-view based and point-based methods. \textit{Volumetric-based} methods, such as  VoxNet~\cite{maturana2015voxnet}, map unstructured 3D point clouds into a set of voxels, and then use 3D Convolutional Neural Networks (CNNs) to perform prediction. 
SEGCloud~\cite{tchapmi2017segcloud} first converts a point cloud into coarse voxels, and then uses 3D fully convolutional network to make prediction.

\textit{Multi view-based} methods usually render a group of 2D images by projecting 3D points from different angles, and then apply 2D image processing models to make prediction. MVCNN~\cite{su2015multi} obtains 2D projection images from 12 views, and uses a CNN model to extract view-based features.
GVCNN~\cite{feng2018gvcnn} leverages the relationship between multiple views (8 or 12) by grouping them based on their discrimination scores. Later on, SimpleView~\cite{goyal2021revisiting} showed that projecting a point cloud onto just six orthogonal planes, and processing these projection images through ResNet~\cite{he2016deep} can be highly effective. More recently, Chen et al.~\cite{chen2023viewnet} introduced ViewNet, which is a two-branch multi-view-based backbone for few-shot 3D classification. 

\textit{Point-based} methods directly take raw point cloud as input.~PointNet~\cite{qi2017pointnet} is a pioneering work in this category, which uses shared MLPs to extract point features, and then applies max-pooling to obtain permutation-invariant features. However, PointNet is unable to capture local features. To address this issue, PointNet++~\cite{qi2017pointnet++} integrates hierarchical sampling, local feature learning from the neighborhood of each point, and multi-scale aggregation to improve point cloud representation. DGCNN~\cite{wang2019dynamic} enhances point cloud processing by dynamically constructing graphs for each point cloud and using EdgeConv layers to capture local geometric features. There are also some multi-modal approaches proposed to fully describe a 3D shape. FusionNet~\cite{hegde2016fusionnet} integrates voxelized data and multi-view images for 3D object classification. CMFF~\cite{yang2023cross} employs DGCNN and ResNet to fuse 3D point cloud data and 2D projection images for few-shot 3D point cloud classification. POV~\cite{ren2024pointofview} is another more recent work utilizing both 2D projection images and 3D point clouds.

\subsection{Domain Adaptation and Generalization for Point Clouds}
Despite the success of supervised 3D point cloud analysis methods,
representing 3D point clouds across different domains still remains as a challenge. Lately, unsupervised domain adaptation (UDA) for point clouds has attracted attention. PointDAN~\cite{qin2019pointdan} is a pioneering work in 3D UDA, aligning local features between source and target data.
DefRec + PCM~\cite{achituve2021self} adopts self-supervised learning. 
It reconstructs partially distorted point clouds and employs mixup to align features across domains. GAST~\cite{zou2021geometry} enhances feature alignment by predicting the rotation angle of a mixed point cloud object and identifying the distorted part of a point cloud. GLRV~\cite{fan2022self} proposed two self-supervised methods, scale prediction and 3D-2D-3D projection reconstruction, to align the distribution of source and target features.

Compared to 3D UDA, 3D domain generalization (DG) is a more realistic, yet more challenging task, since target data is inaccessible during training. Metasets~\cite{huang2021metasets} designs three point augmentation tasks to simulate real-world point data, and proposes to meta-learn representations from a group of augmented point clouds. PDG~\cite{wei2022learning} learns generalizable part-level features with the augmented point clouds as in~\cite{huang2021metasets}.
SUG~\cite{huang2023sug} first splits one source domain into different sub-domains, and proposes a multi-grained sub-domain alignment method to learn the domain-agnostic features. Push-and-pull~\cite{xu2024push} improves the generalization performance by designing a learnable 3D data augmentor to push the augmented samples away from the originals, and a representation regularization objective to pull the predictions back to their origin space. The aforementioned works mainly use DGCNN~\cite{wang2019dynamic} or PointNet~\cite{qi2017pointnet}, to extract point features, which are not the most suitable backbones for DG, as we show below. 

\section{Motivation}
\subsection{Point Utilization Analysis}
\begin{figure}[t!]
\vspace{-0.2cm}
\begin{center}
\includegraphics[width=0.8\linewidth]{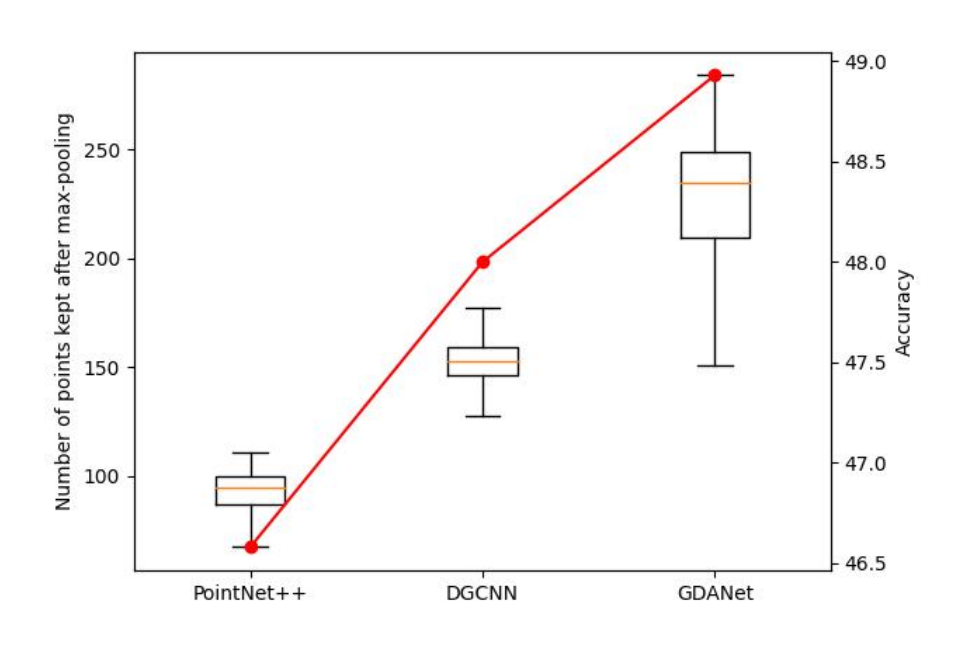}
\end{center}
\vspace{-0.7cm}
    \caption{Classification accuracy versus the number of points retained after max-pooling for PointNet++, DGCNN, and GDANet on PointDA-10.}
\label{fig:point_util}
\vspace{-0.6cm}
\end{figure}

Many point-based methods employ max-pooling, to obtain permutation-invariant features, which causes only a portion of points' features to be used while discarding the rest.~It was shown in~\cite{chen2022discard} that these discarded points are actually useful for supervised learning. In this section, we investigate the number of points utilized after max-pooling, for the DG task on PointDA-10~\cite{qin2019pointdan}, for three point-based methods, namely PointNet++~\cite{qi2017pointnet++}, DGCNN~\cite{wang2019dynamic} and GDANet~\cite{xu2021learning}. PointDA-10 contains three domains: ModelNet~\cite{wu20153d}, ShapeNet~\cite{chang2015shapenet} and ScanNet~\cite{dai2017scannet}. Each subset contains the same ten classes.~ModelNet and ShapeNet are synthetic datasets with points sampled from CAD models, while ScanNet is collected by a real-world scanning sensor. We conduct experiments in two synthetic-to-real scenarios: ModelNet to ScanNet and ShapeNet to ScanNet. Each input point cloud contains 1,024 points. The classification accuracy, shown in Fig.~\ref{fig:point_util}, represent the average values across both settings.~It can be seen that the prediction accuracy is positively correlated with the number of points kept after max-pooling, i.e. point utilization. 
This indicates that models with stronger generalization ability tend to utilize a greater number of points. For example, GDANet achieves the highest accuracy among the three methods, and also has the highest point utilization, with around 221 points, compared to 95 for PointNet++ and 151 for DGCNN. Yet, discarding points is inevitable with max-pooling.~Thus, motivated by this and different from previous works, such as SUG~\cite{huang2023sug} and PDG~\cite{wei2022learning}, we replace the point-based backbone with an alternative backbone to extract more representative features.

\subsection{Multi-View Depth Image Analysis}
\begin{figure}[t!]
\vspace{-0.2cm}
\begin{center}
\includegraphics[width=1\linewidth]{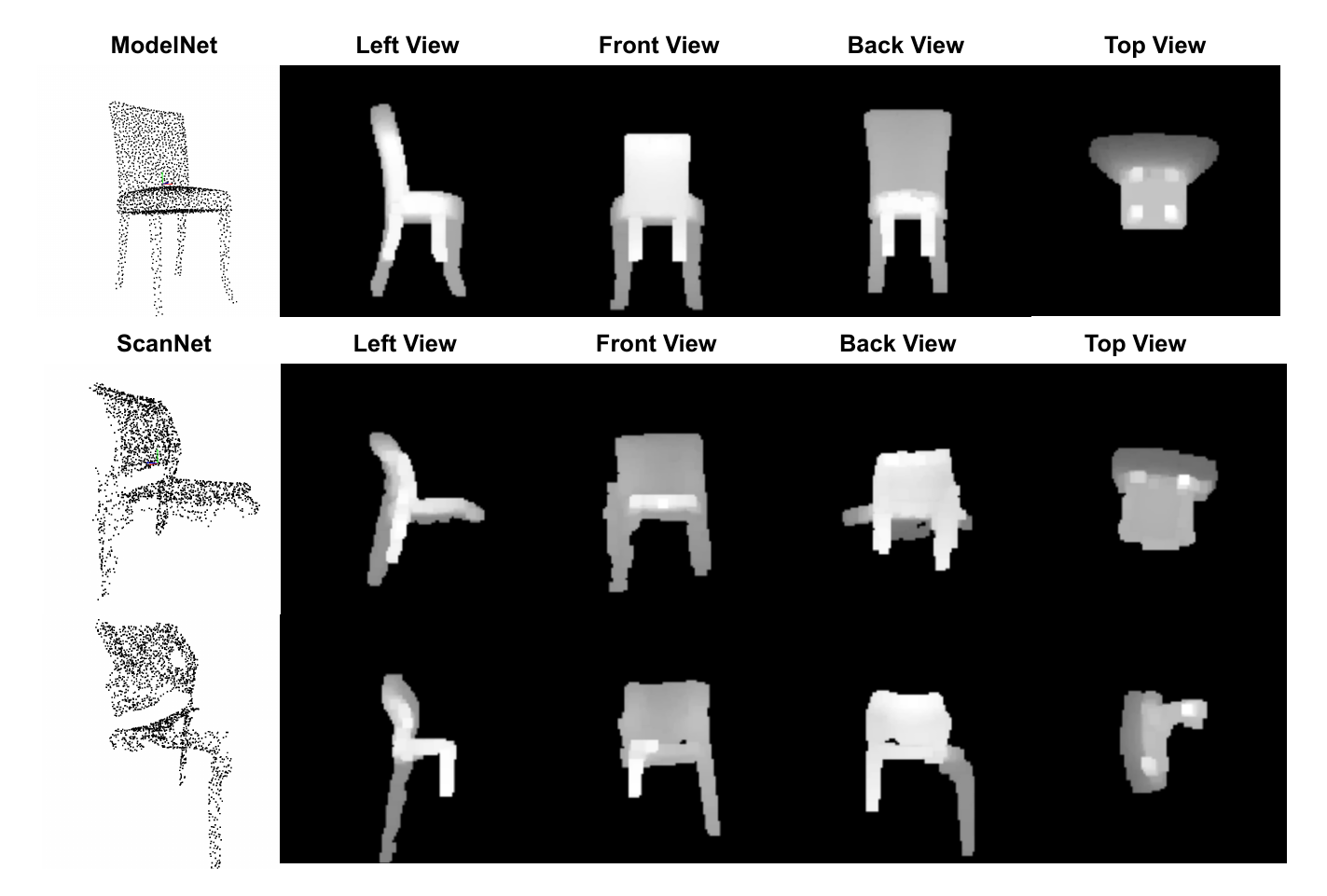}
\end{center}
\vspace{-0.5cm}
    \caption{Example point clouds for chairs and their corresponding depth images for ModelNet and ScanNet datasets.}
\label{fig:projection}
\vspace{-0.5cm}
\end{figure}

As discussed above, when point clouds sampled from CAD models are used as the source domain, and point clouds captured in real-world, by LiDAR sensors, are used as the target domain, significant domain shift occurs. 
Since point-based methods take raw 3D points as input directly, this negatively affects models' generalization ability when dealing with missing points and shape distortions in the target domain. On the other hand, when point clouds are projected into depth images from different angles, certain depth images tend to be more robust against missing points, as shown in Fig.~\ref{fig:projection}, where  the first row and last two rows show example point clouds and their corresponding depth images from ModelNet (source) and ScanNet (target) datasets, respectively.
In ScanNet, the first chair is missing its front legs, 
while the second chair is missing left part of the seat and the right back leg. Despite varying locations of the missing points, some projections, for instance the front and top views of the first ScanNet chair, and the left, front and back views of the second ScanNet chair, still resemble those of the CAD-based point cloud, providing robustness to the issue.~Thus, employing multiple depth images holds more promise for DG compared to using only raw point clouds.  

In this paper, motivated by our findings and observations in terms of point utilization, and depth images, we propose a method that does not rely on point-based backbones (which use max-pooling), and exploits benefits of using multiple depth images instead.


\section{Proposed Method}
\subsection{Problem Definition}
In domain generalization (DG) on point clouds, given a source domain $S=\{x_i,y_i\}_{i=1}^{n_s}$, with $n_s$ labeled synthetic point clouds, our goal is to train a model $f$ on the source domain, which can generalize to an unseen target domain $T=\{x_i,y_i\}_{i=1}^{n_t}$, where $x_i$ and $y_i$ denote the $i^{th}$ point cloud sample and its label in the respective dataset. It should be noted that only the data and labels from the source domain are accessible during training, while data and labels from the target domain are only used for evaluation.

\subsection{Network Architecture}

\begin{figure*}[t!]
\vspace{-0.2cm}
\begin{center}
\includegraphics[width=1\linewidth]{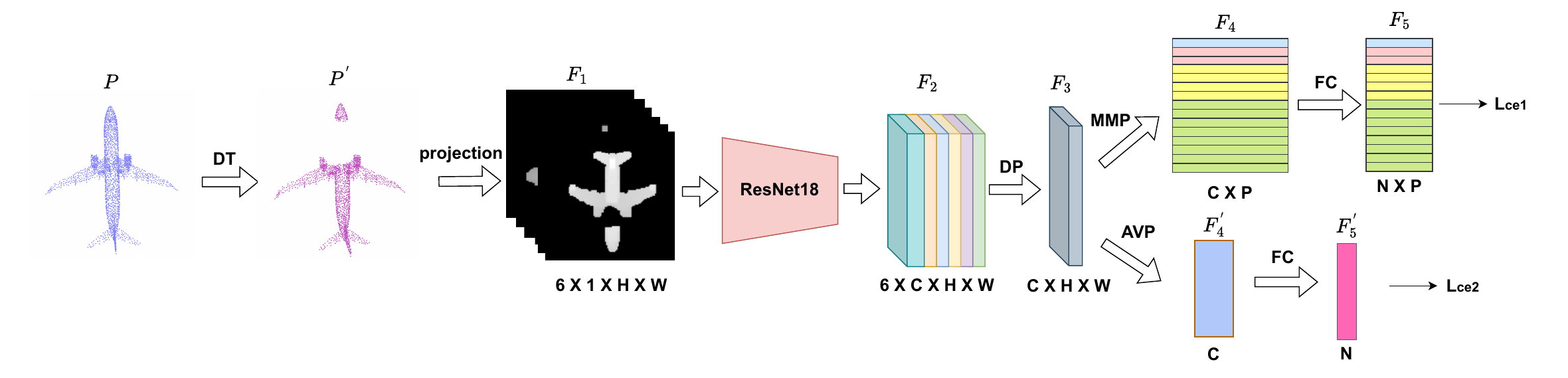}
\end{center}
\vspace{-0.5cm}
    \caption{The pipeline of DG-MVP. $DT$, $DP$, $MMP$, $AVP$ and $FC$ represent Data Transformation, Depth Pooling, Multi-scale Max Pooling, Average Pooling and Fully Connected Layer, respectively. $H$, $W$, $C$, $P$ and $N$ denote height, weight, the number of channels, the total number of strips and the number of class, respectively.}
\label{fig:architecture}
\vspace{-0.5cm}
\end{figure*}

\begin{figure}[t!]
\begin{center}
\includegraphics[width=1\linewidth]{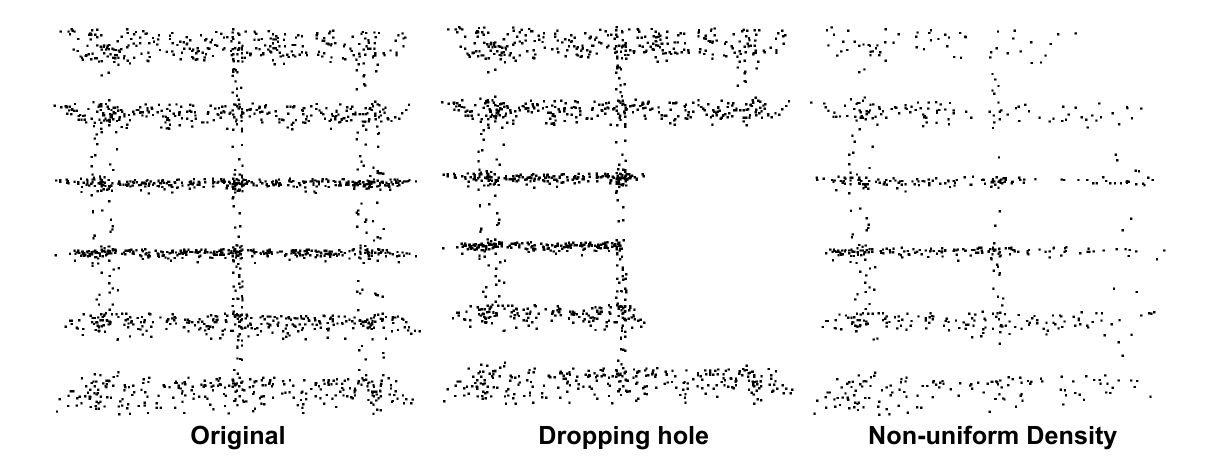}
\end{center}
\vspace{-0.5cm}
    \caption{Examples of augmented point clouds via creating a hole and non-uniform point density.}
\label{fig:augment_pc} 
\vspace{-0.5cm}
\end{figure}

Using the insights gained from the above analysis, we propose a simple but effective DG network, referred to as the DG-MVP, for 3D point cloud classification. We employ multiple projected 2D depth images of a point cloud as input, and adopt ResNet18 as the backbone to extract 2D features. To preserve more global and local features, a multi-scale max pooling module (MMP) is designed and integrated at the end of ResNet18, enhancing the network's ability to capture more representative information.~The overall architecture of our proposed DG-MVP is presented in Fig.~\ref{fig:architecture}. Let $P \in \mathbb{R}^{M\times 3}$ denote an original normalized point set containing $M$ points with their 3D coordinates. To improve the generalization ability of the model, we apply a set of transformations $\{t_i\}_{i=1}^{7}$, as proposed in~\cite{huang2021metasets}, to the original point clouds, simulating the distribution of real-world data. To simulate missing points and occlusion, we randomly select a point $p_i$ and drop its nearest neighbors using various drop rates (0.24, 0.36, 0.45). Density of real-world point clouds is always non-uniform, since it depends how close or far the scanning device is from the object. 
To simulate this scenario, Huang et al.~\cite{huang2021metasets} proposed a method that randomly removes points based on their distance from a randomly selected point on a unit sphere, using a probabilistic approach controlled by a parameter $g$. In our work, we set $g$ to 1.3, 1.4, and 1.6. Examples of augmented point clouds are presented in Fig.~\ref{fig:augment_pc}. Each point cloud is assigned one of the seven possible states: keeping the original cloud or applying one of the six possible random transformation choices (creating holes with three different rates and density adjustment with three different rates). After obtaining the transformed point cloud $P' \in \mathbb{R}^{M'\times 3}$, we project $P'$ onto six orthogonal views~\cite{goyal2021revisiting} to obtain the rendered 2D image $F_1 \in \mathbb{R}^{6\times 1\times H\times W}$, where $6$, $1$, $H$ and $W$ denote the number of projected images and channel, and the  height and width of projection depth image, respectively. Some examples projection images are shown in Fig.~\ref{fig:projection}. 

Next, we use ResNet18 as the feature extractor to obtain feature maps for each depth image, denoted as $F_2 \in \mathbb{R}^{6 \times C \times H \times W}$, where $C$ represents the number of channels, which is 256 in our work. To obtain holistic understanding of the entire object and by drawing inspiration from GaitBase~\cite{fan2023opengait}, which is a gait recognition network,  we use element-wise maximum operation across six views, referred to as the Depth Pooling. Alternatives to Depth Pooling are 
average pooling and view pooling~\cite{chen2023viewnet}. Average pooling computes the mean value across all views. View pooling 
groups the views into pairs (left-right, top-bottom, front-back) and triplets (top-front-left, bottom-back-right), and applies max-pooling along the dimension of the number of projection. Then, these features are concatenated, and fed into a convolutional layer. This module has proven effective for few-shot learning in domain-specific setting. However, these two methods do not perform as well as Depth Pooling for DG. Average pooling can result in the loss of detailed information, leading to inaccurate representation. Moreover, while ModelNet and ShapeNet are axis-aligned, ScanNet is not. 
For instance, in ModelNet, a chair might always be oriented so that its legs are parallel to the vertical axis (z-axis), and the seat is parallel to the horizontal axes (x-axis and y-axis). In ScanNet, the same chair might appear tilted, rotated, or at varying angle. As a result, different views of the same object in the target domain are inconsistent, unlike in the source domain, where views are similar. Therefore, average pooling and view pooling ended up to be ineffective in cross-domain settings, and simple max-pooling yields better results, formulated as $F_3 = maxP(F_2,dim=0)$.
Supporting experimental results are provided under Ablation Studies.

After obtaining 
$F_3 \in \mathbb{R}^{C\times H\times W}$, the model splits into two branches. The bottom branch in Fig.~\ref{fig:architecture} follows the original ResNet architecture, consisting of an average pooling layer and a fully connected (FC) layer. The top branch includes our proposed Multi-scale Max-pooling (MMP) module to capture additional global and local features, followed by an FC layer for prediction. Finally, cross-entropy loss is used in each branch to obtain the losses $L_{ce1}$ and $L_{ce2}$. The overall loss function is:
\begin{equation}
\small{
    L = \lambda_1 L_{ce1} + \lambda_2 L_{ce2},}
    \vspace{-0.2cm}
\end{equation}
where $\lambda_1$ and $\lambda_2$ are hyper-parameters to balance the two loss components.
\vspace{-0.9cm}
\subsection{Multi-scale Max Pooling}
\begin{figure}[t!]
\begin{center}
\includegraphics[width=0.9\linewidth]{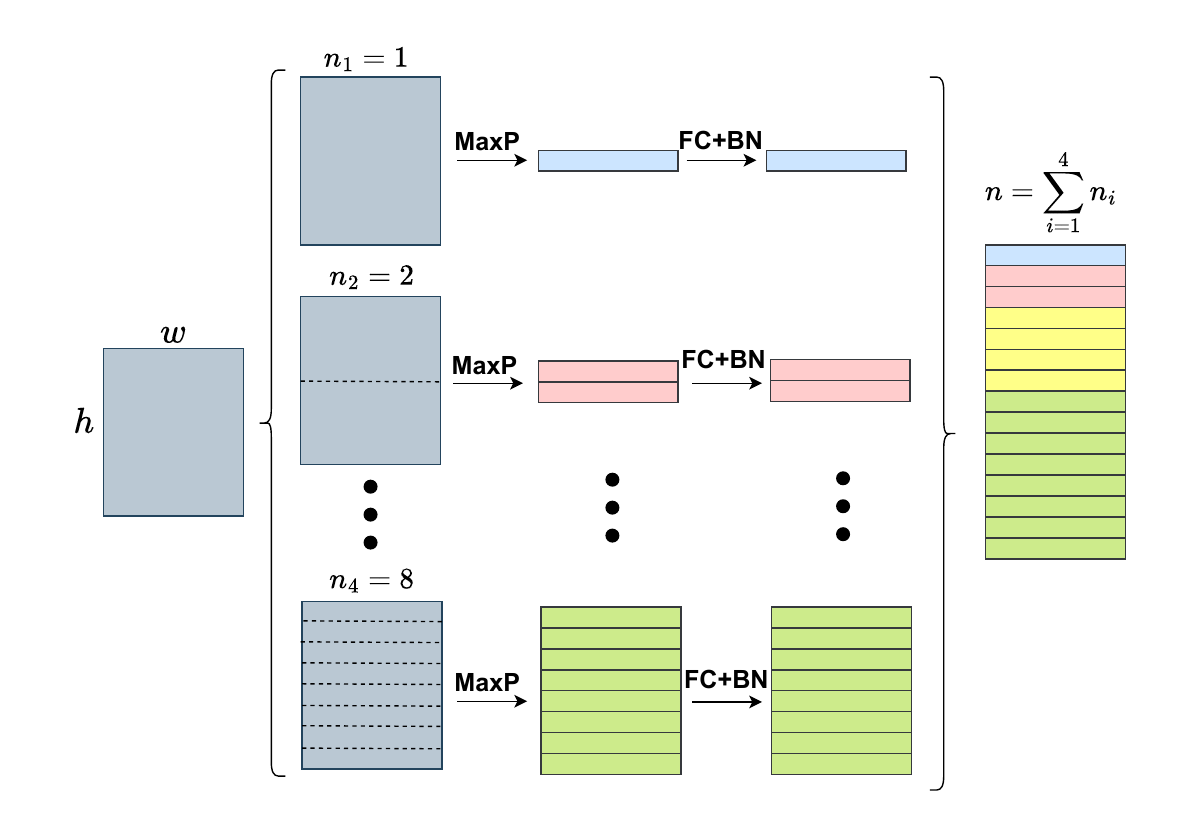}
\end{center}
\vspace{-0.5cm}
    \caption{The structure of Multi-scale Max Pooling. $h$, $w$ and $n_i$ represent the height and weight of the feature map, and the number of strips, respectively. $MaxP$, $FC$ and $BN$ denote max-pooling, fully connected layer and batch normalization, respectively.}
\label{fig:mmp} 
\vspace{-0.5cm}
\end{figure}

We design a modified Multi-scale Max Pooling (MMP), which was originally presented in~\cite{fu2019horizontal} for person re-identification, to capture both broad and fine-grained features. The structure of MMP is shown in Fig.~\ref{fig:mmp}.~The input is $F_3$, which is the output of the Depth Pooling layer. MMP employs different scales: $n_1 =1$, $n_2 =2$, $n_3 =4$ and $n_4 =8$. The input feature map is split into $n_i$ strips along the height dimension, where each strip contains ($\frac{h}{n_i} \times w$)-many pixels. Then, global max-pooling is performed on each strip to get 1-D features. 
Next, an FC layer and batch normalization are applied to each strip to map the features into a discriminative space. It is important to note that each strip has its own FC layer, as strips from different scales represent features with varying receptive fields. Finally, all strips are concatenated to obtain $\sum_{i=1}^{4}n_i$ strips in total.

\section{Experiments}
\subsection{Datasets and Implementation Details}
We evaluate the generalization performance of our proposed DG-MVP on the PointDA-10 and Sim2Real benchmarks. PointDA-10~\cite{qin2019pointdan} is composed of three widely-used datasets: ModelNet ($M$), ShapeNet ($S$), and ScanNet ($S$*), all of which include the same ten categories.~ModelNet contains 4183 training and 856 test
samples.~ShapeNet contains 17378 training and 2492 test samples. ModelNet and ShapeNet are generated by uniformly sampling points from a synthetically generated 3D CAD model. 
Unlike these synthetic datasets, ScanNet is composed of data collected by RGB-D scans in real-world. It contains 6,110 training and 2,048 test samples, with the point clouds usually being incomplete, not axis-aligned, and some point clouds not fully depicting objects. Different from previous works~\cite{xu2024push, huang2023sug}, we conduct experiments solely on the most challenging and realistic scenario, by using synthetic data (ModelNet and ShapeNet) as the source domain and the real-world data (ScanNet) as the target domain. 
For a commensurate comparison with previous works, we follow the same data preparation and experiment settings as in~\cite{zou2021geometry, wei2022learning}. Specifically, each point cloud is down-sampled to 1024 points, and randomly rotated along the z-axis, as described in~\cite{achituve2021self}, for training.

Sim-to-Real~\cite{huang2021metasets} is a 3D DG dataset consisting of three domains: ModelNet ($M$), ShapeNet ($S$) and ScanObjectNN ($SO$)~\cite{uy2019revisiting}.~ScanobjectNN is a real-world point cloud dataset built from scanned indoor scenes. It also contains different deformation variants, such as background occurrence ($SO_B$).~Following~\cite{huang2021metasets}, we consider four synthetic-to-real settings: $M\!\!\rightarrow\!\!SO$ and $M\!\!\!\rightarrow\!\!SO_B$ refer to training on ModelNet and evaluating on ScanobjectNN and ScanObjectNN with background, respectively, with 11 shared classes; $S \rightarrow\!\!SO$ and $S\rightarrow\!\!SO_B$ represent ShapeNet as the source domain, with ScanObjectNN and ScanObjectNN with background as the target domains, sharing 9 classes. We train the models for 200 epochs on PointDA-10 and 160 epochs on Sim-to-Real, with a batch size of 32, using one Quadro RTX 6000 GPU. The Adam optimizer is used with a learning rate of 0.001 and weight decay of 0.00005. Since data transformation is made randomly, we repeat each experiment three times and report the average accuracy.
 %
 %
 %
\subsection{Results on PointDA-10}
We compare our proposed DG-MVP with the SOTA 3D DG methods, including the meta-learning approach MetaSets~\cite{huang2021metasets}, the part-level feature alignment method PDG~\cite{wei2022learning}, the adversarial training method Push-and-Pull~\cite{xu2024push}, and the domain alignment method SUG~\cite{huang2023sug}. We also compare with the SOTA 3D UDA methods, which \emph{utilize target domain data during training}, including PointDAN~\cite{qin2019pointdan}, RS~\cite{sauder2019self}, DefRec + PCM~\cite{achituve2021self}, GAST~\cite{zou2021geometry}, GLRV~\cite{fan2022self} and PC-Adapter~\cite{park2023pc}. 

The results are summarized in Tab.~\ref{tab:pointDA}, where we also report our results without data transformation as ``DG-MVP w/o DT".~We can see that our DG-MVP achieves an average accuracy of 55.29\% in synthetic-to-real scenario, surpassing all other 3D DG methods. It outperforms the second-best model, SUG, by 0.59\% in average accuracy and by 3.71\% in the $S \rightarrow S^*$ scenario. It is also notable that our method even outperforms half of the 3D UDA methods, namely PointDAN, RS and DefRec+PCM, without requiring access to target data during training. These results demonstrate the effectiveness of our proposed DG-MVP. 

\begin{figure}[t!]
\begin{center}
\includegraphics[width=1\linewidth]{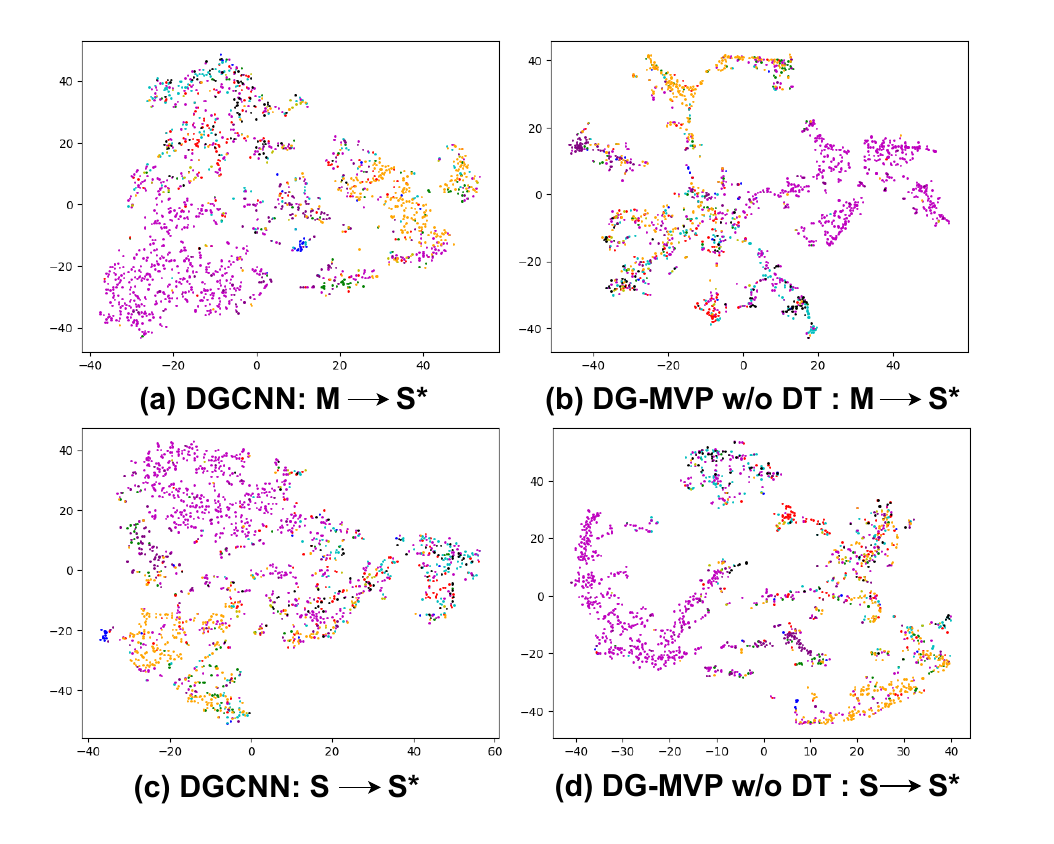}
\end{center}
\vspace{-0.5cm}
    \caption{The t-SNE visualization of feature distribution of the target domain samples on PointDA-10.}
\label{fig:t-sne}
\vspace{-0.2cm}
\end{figure}

All the baselines in Tab.~\ref{tab:pointDA} are point-based and all but one employ DGCNN as their backbone. Thus, we also utilize t-SNE, and visualize the feature distribution on the target domain for DGCNN, and our proposed DG-MVP without data transformation (DG-MVP w/o DT) in Fig.~\ref{fig:t-sne}. It can be observed that the feature distribution of our method is better separated and more distinct compared to DGCNN for both $M \rightarrow S^*$ and $S \rightarrow S^*$ scenarios. This further demonstrates that our DG-MVP can extract better domain-invariant and distinguishing features.

\begin{table}[]
\centering
\resizebox{1\linewidth}{!}{
\begin{tabular}{|l|c|ccc|}
\hline
Method        & UDA/DG               & $M \rightarrow S^*$ & $S \rightarrow S^*$ & Avg            \\ \hline
RS~\cite{sauder2019self}            &                      & 44.8               & 45.7               & 45.25          \\ 
PointDAN~\cite{qin2019pointdan}     & \multirow{5}{*}{UDA} & 45.3               & 46.9               & 46.10         \\
DefRec+PCM~\cite{achituve2021self}    &                      & 51.8               & 54.5               & 53.15          \\ 

PC-Adapter~\cite{park2023pc}    &                      & 58.2               & 53.7               & 55.95          \\    

GAST~\cite{zou2021geometry}          &                      & 59.8               & 56.7               & 58.25          \\   
GLRV~\cite{fan2022self}         &                      & 60.4               & 57.7               & 59.05          \\   

\hline \hline
 
MetaSets~\cite{huang2021metasets}     & \multirow{6}{*}{DG}  & 52.3               & 42.1               & 47.20          \\  
Push-and-Pull~\cite{xu2024push} &                      & 55.3               & 47.0                 & 51.15          \\   
PDG~\cite{xu2021learning}          &                      & \textbf{57.9}      & 50.0                 & 53.95          \\  
SUG~\cite{huang2023sug}          &                      & 57.2               & 52.2               & 54.70          \\     
DG-MVP w/o DT (ours) &               & 50.31               & 51.03               & 52.38          \\ 
DG-MVP (ours) &                      & 54.66              & \textbf{55.91}     & \textbf{55.29} \\ \hline
\end{tabular}}
\vspace{-0.2cm}
\caption{Classification accuracy (in \%) of various 3D UDA and 3D DG methods on PointDA-10 dataset.}
\label{tab:pointDA}
\vspace{-0.3cm}
\end{table}


\begin{table}[tb!]
\centering
\resizebox{0.95\linewidth}{!}{
\begin{tabular}{|c|c|c|c|}

\hline
Method              & $M \rightarrow SO$ & $M \rightarrow SO_B$ & Avg             \\ \hline
PointNet~\cite{qi2017pointnet}            & 58.68              & 52.68               & 55.68           \\ 
DGCNN~\cite{wang2019dynamic}               & 61.37              & 53.26               & 57.32          \\ 
\hline
Metasets~\cite{huang2021metasets} & 60.30               & 62.53                & 61.41           \\ 
PDG~\cite{wei2022learning} & 64.81               & 63.99               & 64.40           \\ \hline
DG-MVP w/o DT (ours)              & 64.63     & 60.79      & 62.71 \\ 
DG-MVP (ours)              & \textbf{70.92}     & \textbf{66.74}      & \textbf{68.83} \\ \hline
\end{tabular}}
\vspace{-0.2cm}
\caption{Accuracy on ModelNet ($M$) to ScanObjectNN w/o background ($SO$) and ScanObjectNN w/ background ($SO_B$) benchmark.}
\label{tab:sim2real_m2s}
\vspace{-0.2cm}
\end{table}
\begin{table}[h]
\centering
\resizebox{0.9\linewidth}{!}{
\begin{tabular}{|c|c|c|c|}
\hline
Method   & $S \rightarrow SO$ & $S \rightarrow SO_B$ & Avg            \\ \hline
PointNet~\cite{qi2017pointnet} & 54.63              & 49.25               & 51.94          \\ 
DGCNN~\cite{wang2019dynamic}    & 54.06              & 51.56               & 52.81          \\ \hline
Metasets~\cite{huang2021metasets} & 52.75              & 54.5                & 53.63          \\ 
PDG~\cite{wei2022learning} & 58.93               & \textbf{56.91}               & 57.92           \\ \hline
DG-MVP w/o DT (ours)              & 56.06     & 52.75      & 54.41 \\ 
DG-MVP (ours)   & \textbf{60.50}     & 56.72      & \textbf{58.61} \\ \hline
\end{tabular}}
\vspace{-0.2cm}
\caption{Accuracy on ShapeNet ($S$) to ScanObjectNN w/o background ($SO$) and ScanObjectNN w/ background ($SO_B$) benchmark.}
\label{tab:sim2real_s2s}
\vspace{-0.6cm}
\end{table}
\subsection{Results on Sim-to-Real}
\begin{figure*}[t!]
\begin{center}
\includegraphics[width=1\linewidth]{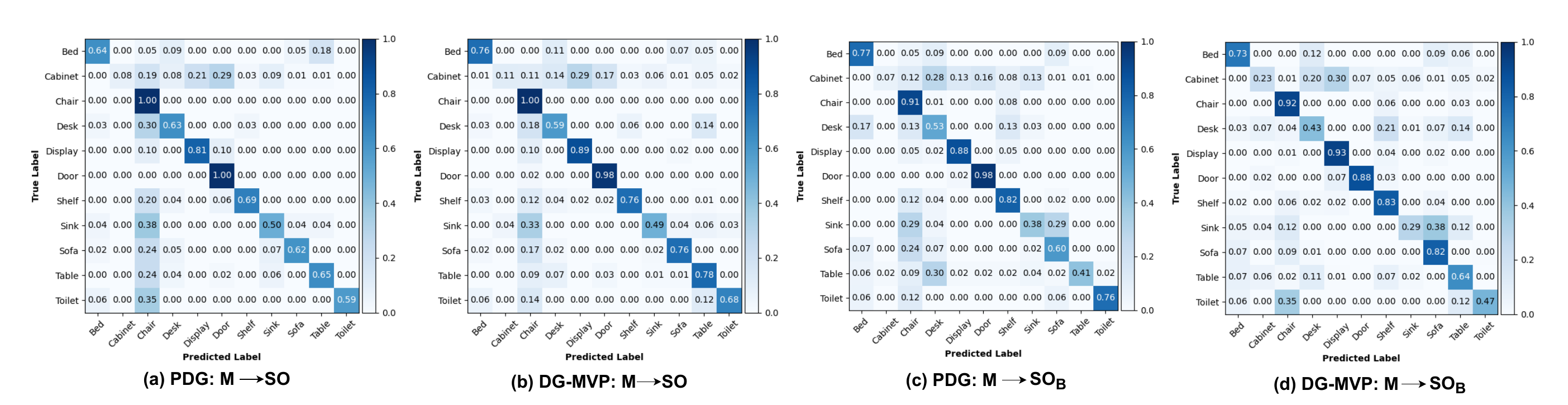}
\end{center}
\vspace{-0.5cm}
    \caption{Confusion matrices for class-wise classification accuracy for PDG and our DG-MVP on ModelNet ($M$) $\rightarrow$ ScanObjectNN w/o background ($SO$) and ModelNet ($M$) $\rightarrow$ ScanObjectNN with background ($SO_B$) settings.}
\label{fig:confusion_matrix}
\vspace{-0.5cm}
\end{figure*}

We conduct experiments on four synthetic-to-real scenarios with this dataset, namely $M\!\!\rightarrow\!\!SO$, $M\!\!\rightarrow\!\!SO_B$, $S\!\!\rightarrow\!\!SO$ and $S\!\!\rightarrow\!\!SO_B$. 
Since the dataset released by~\cite{huang2021metasets} did not include ScanObjectNN with backgrounds, we collected these samples with backgrounds directly from the original ScanObjectNN website. All reported results were obtained by running the published code\footnote{https://github.com/thuml/Metasets/tree/main}\footnote{https://github.com/weixmath/PDG?tab=readme-ov-file}.~We compare with SOTA DG methods, namely Metasets and PDG, which report their results on this dataset. We also compare with point-based backbones used by the baselines in Tab.~\ref{tab:pointDA}, namely PointNet and DGCNN, for reference. The results for overall accuracy are summarized in Tables~\ref{tab:sim2real_m2s} and \ref{tab:sim2real_s2s} when $M$ and $S$ are the source domains, respectively. In Tab.~\ref{tab:sim2real_m2s}, we can see that our method significantly outperforms Metasets and PDG for both $M\!\!\!\rightarrow\!\!SO$ and $M\!\!\!\rightarrow\!\!SO_B$.~On average, it surpasses Metasets and PDG  by 7.42\% and 4.43\%, respectively. 
In addition, to compare with PDG, we show the confusion matrices of class-wise classification accuracy in Fig.~\ref{fig:confusion_matrix}. DG-MVP outperforms PDG by 6.46\% and 0.98\% in terms of average class accuracy in $M\!\!\rightarrow\!\!SO$ and $M\!\!\rightarrow\!\!SO_B$, respectively.
In Tab.~\ref{tab:sim2real_s2s}, our method outperforms Metasets, on average accuracy, with and without data transformation. DG-MVP outperforms PDG by 1.57\% and 0.69\% in the $S \rightarrow SO$ setting and average accuracy, respectively, and achieves comparable results in the $S \rightarrow SO_B$ setting. Moreover, our DG-MVP surpasses, even without data transformation, the most 
commonly used point-based backbone, namely DGCNN, by 5.39\% and 1.6\% in ModelNet to ScanObjectNN and ShapeNet to ScanObjectNN, respectively. These results further validate the effectiveness of our proposed method.

\section{Ablation Studies}
We have conducted ablation studies on the PointDA-10 dataset to study the impact of different components of DG-MVP. In the following, to eliminate the influence of random data transformation choices, each component is evaluated \emph{without applying data transformation}.

\subsection{Comparison of 3D Backbone Architectures}
To verify the effectiveness of our proposed backbone in 3D DG, we compare our method with various point cloud processing backbones, including point-based, projection-based and multi-modality methods. The result summarized in Tab.~\ref{tab:comp_backbone} show that our proposed backbone outperforms all other point cloud processing architectures in terms of average accuracy, while providing significant improvement on ShapeNet to ScanNet ($S\!\!\rightarrow\!\!S^*$). Notably, our DG-MVP even outperforms CMFF~\cite{yang2023cross}, on average and on ($S\!\!\rightarrow\!\!S^*$), which is a multi-modal method that uses DGCNN and ResNet to extract features from both 3D points and multiple depth images, while requiring substantial computational resources. 

\begin{table}[h]
\centering
\resizebox{1\linewidth}{!}{
\begin{tabular}{|c|c|c|c|c|}
\hline
Method    & Input  & $M \rightarrow S^*$ & $S \rightarrow S^*$ & Avg  \\ \hline
PointNet++~\cite{qi2017pointnet++} & Point      & 49.92                    & 43.24                    & 46.58 \\ \hline
DGCNN~\cite{wang2019dynamic}\    & Point      & 50.08                    & 45.92                   & 48.00 \\ \hline
PointCNN~\cite{li2018pointcnn}   & point      & 50.54                    & 45.51                    & 48.03 \\ \hline
GDANet~\cite{xu2021learning}     & Point      & 49.8                     & 48.05                    & 48.93 \\ \hline
ViewNet~\cite{chen2023viewnet}    & Multi-View & 42.85                    & 40.42                    & 41.06 \\ \hline
SimpleView~\cite{goyal2021revisiting} & Multi-View & 49.29                    & 45.22                    & 47.26 \\ \hline
CMFF~\cite{yang2023cross}    & Multi-View + Point & \textbf{50.76}                  & 48.33                    & 49.58 \\ \hline
DG-MVP   & Multi-View & 50.31                   & \textbf{51.03}                    & \textbf{50.67} \\ \hline
\end{tabular}}
\vspace{-0.2cm}
\caption{Comparison of different 3D backbone architectures.}
\label{tab:comp_backbone}
\vspace{-0.5cm}
\end{table}

\subsection{The Impact of Multi-scale Max-pooling}

As explained above, we designed a Multi-scale Max-Pooling (MMP) module to aggregate global and local features. Tab.~\ref{tab:abla_mmp} shows the accuracy of DG-MVP with and without the MMP module, where the numbers inside and outside brackets represent the class average accuracy and overall accuracy, respectively. As can be seen, with MMP, all accuracy values are increased for both $M \rightarrow S^*$ and $S \rightarrow S^*$ settings. For $S \rightarrow S^*$ setting, the overall accuracy (OA) and average class accuracy (ACA) increase as much as 3.73\% and 3.16\%, respectively. 
\begin{table}[h]
\vspace{-0.2cm}
\centering
\resizebox{0.8\linewidth}{!}{
\begin{tabular}{|c|c|c|}
\hline
               & $M \rightarrow S^*$ & $S \rightarrow S^*$ \\ 
               &  OA (ACA) &   OA (ACA) \\\hline
DG-MVP w/o MMP & 49.07 (45.00)         & 47.88 (34.04)      \\ \hline
DG-MVP w MMP   & \textbf{50.31 (45.17)}     & \textbf{51.61 (37.20)}       \\ \hline
\end{tabular}}
\vspace{-0.2cm}
\caption{Average class accuracy and overall accuracy, with and without the MMP for $M \rightarrow S^*$ and $S \rightarrow S^*$.}
\label{tab:abla_mmp}
\vspace{-0.4cm}
\end{table}

\subsection{The Impact of Depth Pooling}
In this study, we analyze the role of Depth Pooling by replacing it with Average Pooling and View Pooling, and comparing the performances. 
The results are summarized in Tab.~\ref{tab:abla_dp}, which shows that Depth Pooling significantly outperforms the other two pooling methods in terms of both average class accuracy and overall accuracy. While View-Pooling captures the relationships between different view angles, it is not effective for domain generalization because real-world data is not axis-aligned. 
\begin{table}[ht!]
\centering
\vspace{-0.2cm}
\resizebox{0.8\linewidth}{!}{
\begin{tabular}{|c|c|c|}
\hline
                & $M \rightarrow S^*$ & $S \rightarrow S^*$ \\ 
                 & OA (ACA) &  OA (ACA) \\
                \hline
Average pooling & 43.75 (43.71)      & 48.84 (36.2)       \\ \hline
View Pooling    & 45.56 (42.24)      & 47.48 (33.57)      \\ \hline
Depth Pooling   & \textbf{50.31 (45.17)}      & \textbf{51.61 (37.2)}       \\ \hline
\end{tabular}}
\vspace{-0.2cm}
\caption{Comparison of Depth Pooling, Average Pooling, and View Pooling.}
\label{tab:abla_dp}
\vspace{-0.3cm}
\end{table}
\subsection{The Impact of ResNet18 Backbone}
To explore the impact of backbone depth and complexity on the generalization ability, we replace ResNet18 with ResNet9 and ResNet50. The results in Tab.~\ref{tab:abla_resnet} show that ResNet50 does not improve the generalization ability. On the other hand, if the backbone is too shallow, it fails to learn representative features. ResNet18 performs best in both the $M \rightarrow S^*$ and $S \rightarrow S^*$ scenarios, leading us to employ it as our backbone.
\begin{table}[hb!]
\centering
\resizebox{0.7\linewidth}{!}{
\begin{tabular}{|c|c|c|}
\hline
Backbone & $M \rightarrow S^*$ & $S \rightarrow S^*$ \\ 
 &OA (ACA) &  OA (ACA) \\
 \hline
ResNet9  & 47.97 (43.64)      & 50.05 (37.1)       \\ \hline
ResNet18 & \textbf{50.31 (45.17)}      & \textbf{51.61 (37.2)}       \\ \hline
ResNet50 & 47.31 (44.94)      & 51.29 (36.87)      \\ \hline
\end{tabular}}
\vspace{-0.3cm}
\caption{The effect of backbone depth in DG.}
\label{tab:abla_resnet}
\vspace{-0.5cm}
\end{table}

\subsection{The Impact of the Number of Views}
The results of using different number of views to obtain depth images to train our model are summarized in Tab.~\ref{tab:abla_num_views}. In the case of 6 views, we use the 6 orthogonal views. For 8 views, we select the projections from the vertices of the cube centered on the object. For 14 clock views, we select the first 12 views around the object every 30 degrees. Remaining two views are top and bottom views. For 14 cube views, we combine the 6 orthogonal views and 8 views. The results illustrate that using 6 orthogonal views provides the best average accuracy. While 14 cube views in the $M \rightarrow S^*$ setting and 14 clock views in the $S \rightarrow S^*$ setting perform slightly better (by 0.68\% and 0.17\%, respectively), they do not improve class average accuracy and require \textit{significantly more} computing power. Using 6 views strikes a balance by conserving computational resources while effectively capturing the object's features.
\begin{table}[hb!]
\centering
\resizebox{0.9\linewidth}{!}{
\begin{tabular}{|c|c|c|c|}
\hline
No. of views & $M \rightarrow S^*$ & $S \rightarrow S^*$ & Avg           \\ 
&OA (ACA) &  OA (ACA) &  OA (ACA) \\ \hline
8 views         & 44.66 (40.8)                  & 45.28 (34.45)                  & 44.97 (37.63)             \\ \hline
14 clock views  & 49.80 (41.84)      & \textbf{51.78 (37.06)}      & 50.79 (39.45) \\ \hline
14 cube views   & \textbf{50.99 (43.34)}      & 49.41 (35.48)      & 50.2 (39.41)  \\ \hline
6 views (ours)   & 50.31 (45.17)      & 51.61 (37.2)       & \textbf{50.96} (41.19) \\ \hline
\end{tabular}}
\vspace{-0.2cm}
\caption{The comparison of using different number of views.}
\label{tab:abla_num_views}
\end{table}

\vspace{-0.5cm}
\section{Conclusion}
In this paper, we have proposed a novel and effective domain generalization method, DG-MVP, for 3D point cloud classification. In order to mitigate the domain shift between synthetic and real-world point clouds, caused mainly by missing points and occlusion, we employ six different depth images, obtained by projecting a point cloud onto six orthogonal planes, as input and also apply various data transformations
to simulate real-world data. 
Furthermore, we have employed Depth Pooling to extract the most prominent features across all projection images. To obtain more global and local features, we have proposed a modified Multi-scale Max-Pooling approach , which splits the feature maps into many strips, and performs max-pooling at each strip. We have performed extensive experiments on the PointDA-10 and Sim-to-Real benchmarks for all synthetic-to-real settings. Results have shown that the proposed DG-MVP outperforms existing 3D domain generalization methods in most scenarios and even surpasses several 3D domain adaptation methods, which require access to target domain data during training.  We have also conducted ablation studies and additional analyses on PointDA-10 to validate the effectiveness of the components of our proposed approach.

{\small
\bibliographystyle{ieee_fullname}
\bibliography{main}
}

\end{document}